%% file: cameraready_iccv.tex
\documentclass[10pt,twocolumn,letterpaper]{article}

\usepackage{iccv}              

\input{cvpr_preamble}

\input{settings_cvpr}

\definecolor{iccvblue}{rgb}{0.21,0.49,0.74}
\usepackage[pagebackref,breaklinks,colorlinks,allcolors=iccvblue]{hyperref}

\def\paperID{*****} 
\def\confName{ICCV}
\def\confYear{2025}

\title{Disentangling Static and Dynamic Information\\
for Reducing Static Bias in Action Recognition}

\author{Masato Kobayashi,
Ning Ding, Toru Tamaki\\
Nagoya Institute of Technology\\
Nagoya, Japan}

\begin{document}
\maketitle

\begin{abstract}
    \input{main_abstract}
\end{abstract}

\input{main_text}

\section*{\uppercase{Acknowledgements}}
This work was supported in part by JSPS KAKENHI Grant Number JP22K12090 and 25K03138.

\input{appendix}

{
    \small
    \bibliographystyle{ieeenat_fullname}
    \bibliography{mybib,all}
}

\end{document}

%% file: cvpr_preamble.tex
\usepackage[dvipsnames]{xcolor}


%% file: settings_cvpr.tex
\usepackage{graphicx}
\usepackage{amsmath}
\usepackage{amssymb}

\usepackage{multirow}
\usepackage{multicol}
\usepackage{diagbox}
\usepackage{xfrac}

\usepackage[labelformat=simple,subrefformat=simple]{subcaption}
\usepackage{caption}
\usepackage{algorithm}
\usepackage{algpseudocode}


\usepackage{minted}



%% file: main_abstract.tex
Action recognition models rely excessively on static cues rather than dynamic human motion, which is known as static bias. This bias leads to poor performance in real-world applications and zero-shot action recognition. In this paper, we propose a method to reduce static bias by separating temporal dynamic information from static scene information. Our approach uses a statistical independence loss between biased and unbiased streams, combined with a scene prediction loss. Our experiments demonstrate that this method effectively reduces static bias and confirm the importance of scene prediction loss.

%% file: main_text.tex
\section{Introduction}

Action recognition is the task of recognizing human actions in videos, which is expected to be used in a variety of situations, such as surveillance camera systems and gesture recognition
\cite{Hara_IEICE2020_ActionRecognition,Kong_arXiv2022_ActionRecognitionPredictionSurvey,Ulhaq_arXiv2022_VisionTransformersActionRecognitionSurvey}.

However, action recognition models tend to make predictions based on static cues (such as background or objects) rather than human action \cite{Chung_NEURIPS2022_Action-Swap,Li_ECCV2018_Resound}. This problem is called static bias, a type of representation bias.
In the context of action recognition,
it is called \emph{background bias} when the prediction is strongly influenced by the background.
\emph{Single frame bias} \cite{Lei_ACL2023_RevealingSingleFrameBias,Li_CVPR2024_MVBench}
has also been reported, with which bias
the model is able to predict actions from a single frame of a video.
Models that rely on static bias to predict actions apparently work better, but this is often not the case in real-world applications and zero-shot action recognition, leading to degraded performance.

One of the causes of bias is the spurious correlation between static cues of the scene and temporal information of the dynamics of human actions.
Therefore, prior methods have been proposed to suppress the correlation by modifying or masking the background and foreground \cite{Chung_NEURIPS2022_Action-Swap,Li_ICCV2023_StillMix,Kimata_MMAsia2022_ObjectMix,Wang_CVPR2021_RemovingBackground,Kim_NEURIPS2022_SynAPT,Zhong_NEURIPS2023_PPMA,Sugiura_VISAPP2024_S3Aug,Fukuzawa_MMM2025_Zero_shot}
or adapted model architectures
\cite{Bae_ECCV2024_Devias,Choi_NEURIPS2019_Dance-in-the-mall,Yun_PMLR2022_TIME,Duan_ECCV2022_Omnidebias,Bao_ICCV2021_CED,fioresi_ICLR2025_ALBER}.
However, these approaches widen the gap between training and inference because masking the background severely alters the appearance of the video frames.
In addition, they do not answer the fundamental question: ``How to disentangle action dynamics and static appearance information.''
While there are several studies that disentangle dynamic and static information in videos \cite{Bae_ECCV2024_Devias, Liu_AAAI2021_MAP-IVR},
they do not aim to counteract static bias.

To address this, we propose extracting dynamic features that are statistically independent of static bias, thereby enabling the disentanglement of static and dynamic information obtained from videos.
The proposed architecture has two streams, where the static and temporal features extracted by each stream are push apart from each other, as well as an additional scene classification stream utilizing adversarial losses.
The proposed method does not need preprocessing on the video, such as masking background as in prior work;
hence, feature representation is consistent in training and inference,
leading to a small gap.

\section{Related work}

\subsection{Disentangled representation learning}

Since videos have not only a visual appearance but also temporal dynamics, it is natural to learn disentangled representation separated into independent dynamic and static information, and
some methods have been proposed to disentangle videos into static and dynamic features.
DEVIAS \cite{Bae_ECCV2024_Devias} learns to disentangle videos into dynamic person and static background regions,
while MAP-IVR \cite{Liu_AAAI2021_MAP-IVR} learns to disentangle features based on the similarity of image and video features using orthogonal constraints.

Although these studies aim at disentangling the representation of videos,
it is not clear how the features of the video should be separated into dynamic action information and static appearance information. 
DEVIAS extracts features that represent motion using estimated person masks, which is inherently the same as masking background in the original frames.
MAP-IVR is a model for image-to-video retrieval, and hence the temporal features of videos might not be separated from the appearance features from images.

\begin{figure}[t]
    \centering

    \subcaptionbox{with extractor-based biased stream
        \label{fig:method2_DiffArch_GRL}}{%
        \includegraphics[width=1.0\linewidth]{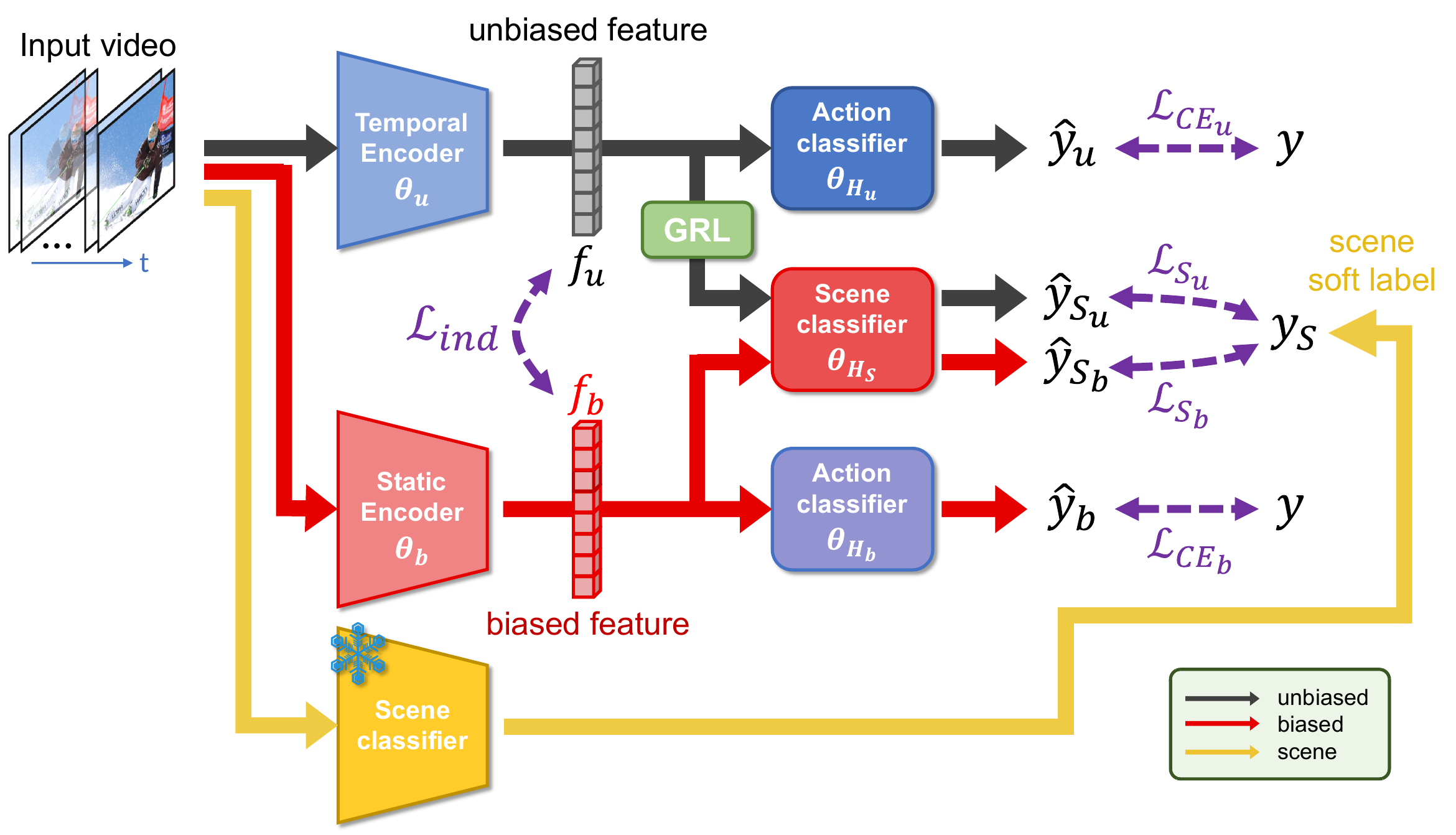}%
    }

    \vspace*{2em}

    \subcaptionbox{with input-based biased stream
        \label{fig:method2_DiffInput_GRL}}{%
        \includegraphics[width=1.0\linewidth]{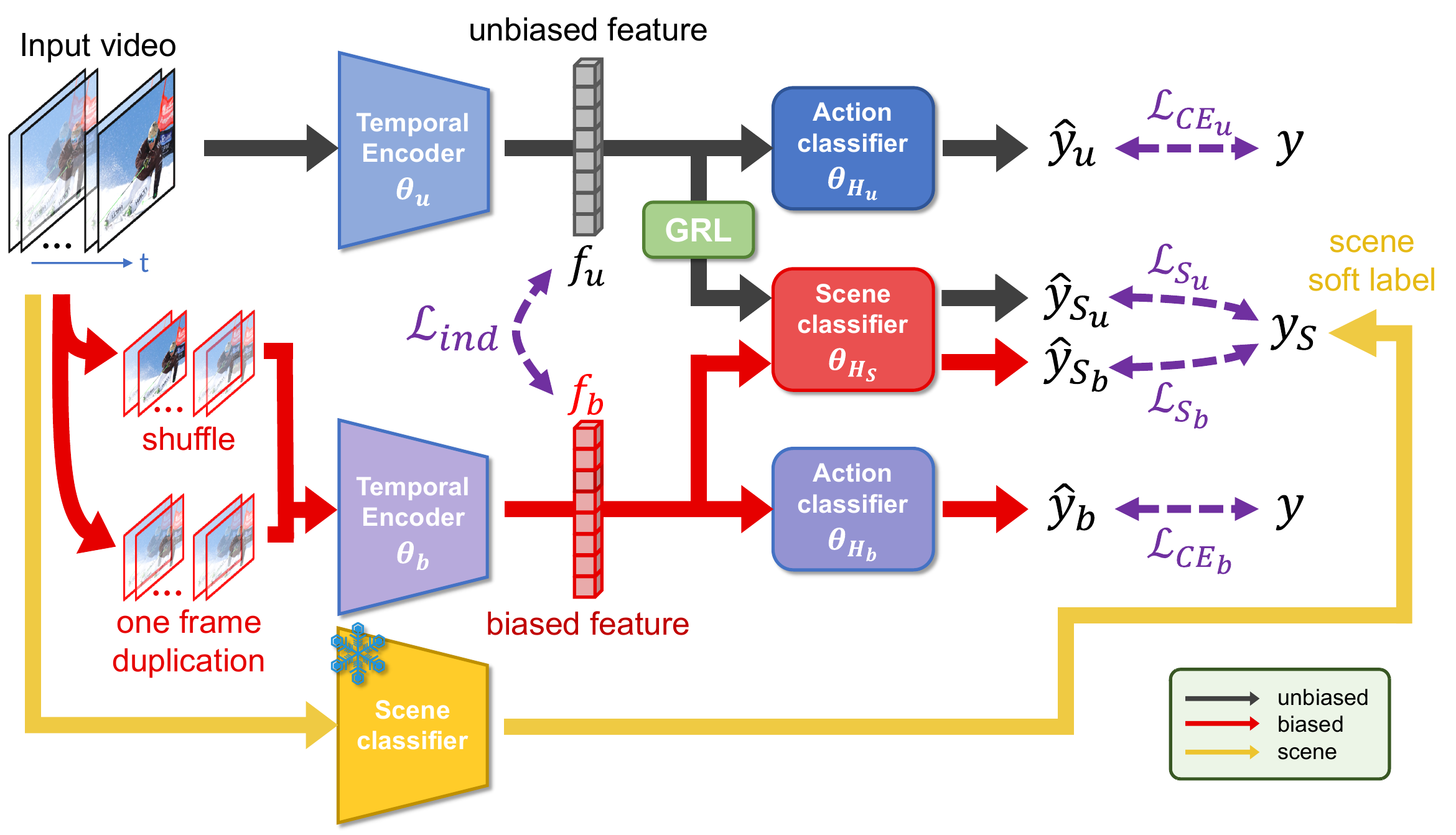}%
    }

    \caption{
        \textbf{Overview of the proposed method.}
        The black path indicates the \emph{unbiased} stream, while the red path indicates the \emph{biased} stream. The yellow path represents a scene-based soft label.
        }
    \label{figs:method2_network_GRL}
\end{figure}

\subsection{Reducing static bias}

Methods for reducing static bias can be roughly divided into two categories; modifying or masking the background or foreground objects in frames of training videos, and adapting model architectures and training strategies.

In the first category, various methods have been proposed;
Action-Swap \cite{Chung_NEURIPS2022_Action-Swap}, 
StillMix \cite{Li_ICCV2023_StillMix}, 
and Background Erasing \cite{Wang_CVPR2021_RemovingBackground}
replace the background with different ones from other videos,
S3Aug \cite{Sugiura_VISAPP2024_S3Aug} generates a different background,
Fukuzawa et al. \cite{Fukuzawa_MMM2025_Zero_shot} simply masked the background,
SynAPT \cite{Kim_NEURIPS2022_SynAPT} and PPMA \cite{Zhong_NEURIPS2023_PPMA}
synthesize the foreground regions of the actor.
However, these methods that modify video frames during training do not clearly specify how to handle frames at inference, resulting in a substantial domain gap between training and inference.

In the second category, many approaches have attempted to disentangle the video representation.
TIME \cite{Yun_PMLR2022_TIME} employs self-supervised learning with the pretext task of predicting the order of the frames.
Choi et al. \cite{Choi_NEURIPS2019_Dance-in-the-mall} proposed
to reduce the confidence in the prediction of actions for videos in which people are masked.
DEVIAS \cite{Bae_ECCV2024_Devias} estimates the mask of the actor regions using a slot attention to mask the background features.
OmniDebias \cite{Duan_ECCV2022_Omnidebias} reweights features of action regions and applies multiple adversarial losses to make scene recognition fail while action recognition success.
CED \cite{Bao_ICCV2021_CED} uses metric learning to extract features that are independent of static biases.
Most of these approaches do not clarify how to disentangle dynamic and static information at the feature level
\cite{Bae_ECCV2024_Devias, Choi_NEURIPS2019_Dance-in-the-mall},
or
do not have any constraints that explicitly disentangle video features into static and dynamic ones
\cite{Duan_ECCV2022_Omnidebias,Yun_PMLR2022_TIME}.

In contrast, Contrastive Evidence Debiasing (CED) \cite{Bao_ICCV2021_CED} reduces static bias by explicitly disentangling dynamic features from video features at the feature level with three streams of 3D CNN.
It uses the Hilbert-Schmidt Independence Criterion (HSIC) \cite{Bahng_ICML2020_Rebias,Song_JMLR2012_DependenceMaximization}, which measures the nonlinear dependence between two variables, 
to learn how to separate the output of two biased streams that only use static bias from the output of the unbiased stream that uses temporal features.
Inspired by CED, we propose two-stream architectures
using recent Transformer models,
enhanced by introducing an adversarial classifier
as in \cite{Duan_ECCV2022_Omnidebias,Choi_NEURIPS2019_Dance-in-the-mall}.

\section{Method}
\label{sec:3}

Figure \ref{figs:method2_network_GRL} shows an overview of the proposed method. We present two versions with a similar architecture.
These models have two streams; one is the \emph{unbiased} stream
that predicts actions based on unbiased feature $f_u$ with temporal dynamics,
and the other is the \emph{biased} stream providing biased feature $f_b$ that is inaccessible to temporal information and hence needs to completely rely on static bias.
We use metric learning with HSIC so that the features of the two streams become statistically independent, in order to make the unbiased stream pay more attention to temporal features.
Furthermore, we add adversarial losses with a scene-based soft label for scene prediction to each stream to ensure that the feature of the unbiased stream has no background information.

For the unbiased stream, we use a commonly used action recognition model
\cite{Bertasius_ICML2021_TimeSformer}
because it is expected to capture dynamic temporal information from videos.
In the following section, we explain the biased stream, scene-based adversarial losses, and metric learning to separate biased and unbiased streams.

\subsection{Biased stream}
\label{sec:3.1}

The concept of the proposed method is to separate the unbiased feature $f_u$ from the biased feature $f_b$ of the biased stream, and we propose two versions, as shown in Figures \ref{fig:method2_DiffArch_GRL} and \ref{fig:method2_DiffInput_GRL}.
In the first version shown in Fig. \ref{fig:method2_DiffArch_GRL} ,
the biased stream (red path) uses a feature extractor for images, applied to each frame of the video followed by a simple temporal average to generate a video-level action prediction. This does not capture any temporal information because the order of the frames does not affect the result.
We call this version \emph{extractor-based biased stream}.

In the second approach
shown in Fig. \ref{fig:method2_DiffInput_GRL}, the feature extractor of the biased stream (red path) is the same as that of the unbiased stream (black path); hence it can capture dynamic information. However, we change the input video for the biased stream so that this stream relies more on the static cues; alternating a shuffled version in which the order of frames of the video are randomly shuffled, and a duplicate version in which all frames of the video are copies of a randomly selected single frame.
This is similar to CED \cite{Bao_ICCV2021_CED}, and is
hereafter called \emph{input-based biased stream}.

\subsection{Adversarial loss through scene prediction}
\label{sec:3.2}

Furthermore, we introduce scene prediction losses to each stream
to further prompt the disentanglement
(yellow path in Fig.\ref{figs:method2_network_GRL}). As a result, biased features incorporate background information for scene prediction,
while unbiased features do not.
To this end, we add a gradient reversal layer (GRL) \cite{Ganin_jMLR2016_GRL} to the unbiased stream only to make it fail for scene prediction, while the biased stream succeeds.

To avoid a costly annotation of scene labels for videos in action recognition datasets, we pretrain a ViT \cite{Dosovitskiy_ICLR2021_ViT_Vision_transformer} on Place365 \cite{Lopez-Cifuentes_arXiv2019_Place365} then freeze it. When training the proposed models, we pick up a random frame of the input video and use the pre-trained and frozen ViT to generate a soft label $y_S$. KL divergence with this soft label is used as the losses $\mathcal{L}_{S_u}$ and $\mathcal{L}_{S_b}$ for the predictions $\hat{y}_{S_u}, \hat{y}_{S_b}$ of the scene by the unbiased branch with GRL and the biased branch.

However, the scene prediction by each branch could be noisy at the beginning of the training.
Therefore, these losses are not applied during the early epochs of training. After reaching epoch $t_0$, we add them and continue the training process.

\subsection{Separating biased and unbiased features}
\label{sec:3.3}

Next, we explain how to extract the unbiased feature $f_u$
independently of the biased feature $f_b$ affected by the static bias.
We formulate a min-max optimization using the Hilbert-Schmidt Independence Criterion (HSIC) \cite{Bahng_ICML2020_Rebias,Song_JMLR2012_DependenceMaximization}
as in previous work \cite{Bao_ICCV2021_CED}.
HSIC provides a quantitative assessment of the level of statistical independence between two multidimensional variables.
It is non-negative, with smaller values indicating a higher degree of independence.
Therefore, minimizing HSIC between biased and unbiased features
increases the statistical independence between two streams.
In each training step with a batch size of $m$ samples, we compute the HSIC between the two sets of biased and unbiased features $f_b$ and $f_u$, each with $m$ samples, as the independence loss $\mathcal{L}_{\mathit{ind}}$.

The final loss $\mathcal{L}$ is a weighted sum of the losses of the two streams, the unbiased stream loss $\mathcal{L}_u$ and the biased stream loss $\mathcal{L}_b$, shown as below;
\begin{align}
    \mathcal{L}_u =&\
    \mathcal{L}_{\mathit{CE}_u}(\hat{y}_u, y; \theta_u, \theta_{H_u}) \nonumber\\
    &+ \beta(t) \, \mathcal{L}_{\mathit{S}_u}(\hat{y}_{S_u}, y_S; \theta_u, \theta_{H_S}) \nonumber\\
    &+ \lambda \, \mathcal{L}_{\mathit{ind}}(f_b,f_u;\theta_u, \mathrm{sg}(\theta_b))
    \label{eq:DiffArch_Lu}
    \\
    \mathcal{L}_b =&\
    \mathcal{L}_{\mathit{CE}_b}(\hat{y}_b, y; \theta_b, \theta_{H_b}) \nonumber\\
    &+ \beta(t) \, \mathcal{L}_{\mathit{S}_b}(\hat{y}_{S_b}, y_S; \theta_b, \theta_{H_S}) \nonumber\\
    &- \lambda \, \mathcal{L}_{\mathit{ind}}(f_b,f_u;\mathrm{sg}(\theta_u), \theta_b) 
    \label{eq:DiffArch_Lb}
    \\
    \mathcal{L} =&\
    \alpha \mathcal{L}_u + (1 - \alpha) \mathcal{L}_b,
    \label{eq:DiffArch_L}
\end{align}
where
$\beta(t) = u(t-t_0) \beta$ where $u(t)$ is a step function that activates after $t_0$ epoch,
$\alpha, \beta$ and $\lambda$ are hyperparameters.
$\mathcal{L}_{\mathit{CE}}$ is the cross-entropy loss with the action label for predictions $\hat{y}_u, \hat{y}_b$ by each stream.
$\theta_b$ and $\theta_u$ are the parameters of the biased and unbiased streams,
$\theta_{H_b}, \theta_{H_b}, \theta_{H_S}$ are the parameters of each head.
The third term of Eq.\eqref{eq:DiffArch_Lu}  encourages $f_u$ to be independent of $f_b$ only by updating $\theta_u$ with a stop gradient (sg) for $\theta_b$. In contrast, the third term of Eq.\eqref{eq:DiffArch_Lb} adversarially updates $f_b$ to approach $f_u$, while gradients flow only for $\theta_b$.

\section{Experiments}
\label{sec:4}

\begin{table*}[t]
    \centering
    
    \caption{
        \textbf{Accuracy and metrics for static bias on Temporal32.}
        The top row is a baseline of a single unbiased stream with TimeSformer.
        HSIC values are expressed in units of 1e-4.
        }
    \label{tab:Temporal32_result}

        \begin{tabular}{c|c|c|cccccc}
          \shortstack{biased\\ stream} & 
          \shortstack{scene\\ prediction} & 
          $t_0$ & top-1
          & BOR $\downarrow$ & HOR $\uparrow$ & SHAcc $\uparrow$ & SBErr $\downarrow$ & HSIC $\downarrow$ \\ \hline
          
          
         --- & --- & ---
          & \textbf{80.9} & 77.0 & 23.4 & 9.61 & 36.3 & --- \\ \hline
          
          
          extractor-based & --- &---
            & 76.8 & 73.5 & \textbf{33.2} & \textbf{16.1} & 33.7 & \textbf{2.74}  \\ \hline
          input-based & --- & ---
            & 74.5 & 70.4 & 31.2 & \textbf{16.1} & 30.7 & 6.34 \\ \hline

          
          \multirow{2}{*}{extractor-based} & \multirow{2}{*}{$\checkmark$} & 0  
            & 75.0 & 74.1 & 28.7 & 15.3 & 29.9 & 5.29 \\
          \multirow{2}{*}{} && 15 
            & 78.0 & 71.7 & 28.4 & 14.4 & 31.8 & 2.81 \\ 
          \hline
          \multirow{2}{*}{input-based} & \multirow{2}{*}{$\checkmark$} & 0  
            & 69.5 & 71.0 & 29.1 & 15.0 & \textbf{27.5} & 6.82 \\
          \multirow{2}{*}{} && 15 
            & 73.9 & \textbf{66.5} & 26.5 & 14.6 & 29.4 & 6.50 \\ \hline

        \end{tabular}
\end{table*}

\begin{table*}[t]
    \centering
    
    \caption{
        \textbf{Accuracy and metrics for static bias on UCF101.}
        HSIC values are expressed in units of 1e-4.
        }
    \label{tab:UCF101_result}
    
        \begin{tabular}{c|c|c|cccccc}
          \shortstack{biased\\ stream} & 
          \shortstack{scene\\ prediction} & 
          $t_0$ & top-1
          & BOR $\downarrow$ & HOR $\uparrow$ & SHAcc $\uparrow$ & SBErr $\downarrow$ & HSIC $\downarrow$ \\ \hline
          
          
          --- & --- & ---
          & \textbf{91.8} & 69.2 & 28.9 & 19.6 & 32.7 & --- \\ \hline
          
          
          extractor-based &--- & --- 
            & 88.9 & \textbf{59.7} & 40.1 & 24.7 & \textbf{24.6} & 5.87  \\ \hline
          input-based &--- & --- 
            & 91.3 & 60.8 & 37.9 & 25.7 & 25.7 & 5.90 \\ \hline

          
          \multirow{2}{*}{extractor-based} & \multirow{2}{*}{$\checkmark$} & 0  
            & 89.4 & 61.0 & 38.0 & 24.5 & 26.7 & 5.34 \\
          \multirow{2}{*}{} && 15 
            & 90.3 & 62.6 & 34.7 & 23.9 & 26.8 & \textbf{5.08} \\ \hline
          \multirow{2}{*}{input-based} & \multirow{2}{*}{$\checkmark$} & 0  
            & 91.0 & 62.5 & \textbf{40.7} & \textbf{26.7} & 26.7 & 6.23 \\
          \multirow{2}{*}{} && 15 
            & 91.6 & 60.9 & 39.1 & 25.5 & 27.5 & 5.90 \\ \hline

        \end{tabular}
\end{table*}

\begin{table*}[t]
    \centering
    
    \caption{
        \textbf{Accuracy and metrics for static bias on UCF-Temporal13.}
        HSIC values are expressed in units of 1e-4.
        }
    \label{tab:UCF-temporal13_result}
    
        \begin{tabular}{c|c|c|cccccc}
          \shortstack{biased\\ stream} & 
          \shortstack{scene\\ prediction} & 
          $t_0$ & top-1
          & BOR $\downarrow$ & HOR $\uparrow$ & SHAcc $\uparrow$ & SBErr $\downarrow$ & HSIC $\downarrow$ \\ \hline
          
          
          --- & --- & ---
          & 80.1 & 79.4 & 37.8 & 23.6 & 31.2 & --- \\ \hline
          
          
          extractor-based &--- & --- 
            & 76.0 & 70.1 & \textbf{65.2} & \textbf{34.4} & 24.3 & 35.8  \\ \hline
          input-based &--- & --- 
            & \textbf{83.8} & \textbf{67.3} & 54.4 & 33.7 & 24.6 & 35.6 \\ \hline

          
          \multirow{2}{*}{extractor-based} & \multirow{2}{*}{$\checkmark$} & 0  
            & 78.4 & 70.5 & 56.0 & 32.5 & 24.8 & 34.9 \\
          \multirow{2}{*}{} && 15 
            & 79.8 & 71.9 & 54.0 & 29.1 & 26.0 & \textbf{31.6} \\ \hline
          \multirow{2}{*}{input-based} & \multirow{2}{*}{$\checkmark$} & 0  
            & 79.2 & 68.8 & 56.3 & 28.4 & \textbf{23.2} & 36.4 \\
          \multirow{2}{*}{} && 15 
            & 80.5 & 71.0 & 61.3 & 33.6 & 27.7 & 35.7 \\ \hline

        \end{tabular}
\end{table*}

\begin{table*}[t]
    \centering
    
    \caption{
        \textbf{Accuracy and metrics for static bias on UCF-Static13.}
        HSIC values are expressed in units of 1e-4.
        }
    \label{tab:UCF-static13_result}
    
        \begin{tabular}{c|c|c|cccccc}
          \shortstack{biased\\ stream} & 
          \shortstack{scene\\ prediction} & 
          $t_0$ & top-1
          & BOR $\downarrow$ & HOR $\uparrow$ & SHAcc $\uparrow$ & SBErr $\downarrow$ & HSIC $\downarrow$ \\ \hline
          
          
          --- & --- & ---
          & 93.0 & 65.8 & 21.8 & 11.7 & 37.8 & --- \\ \hline
          
          
          extractor-based &--- & --- 
            & \textbf{95.6} & \textbf{51.3} & 25.6 & 13.8 & 28.2 & 42.1  \\ \hline
          input-based &--- & --- 
            & 92.5 & 58.5 & 27.0 & 14.7 & \textbf{29.4} & 39.5 \\ \hline

          
          \multirow{2}{*}{extractor-based} & \multirow{2}{*}{$\checkmark$} & 0  
            & 90.4 & 55.5 & 31.0 & 12.5 & 30.4 & 39.6 \\
          \multirow{2}{*}{} && 15 
            & 93.9 & 59.6 & 25.5 & \textbf{15.5} & 34.1 & \textbf{37.3} \\ \hline
          \multirow{2}{*}{input-based} & \multirow{2}{*}{$\checkmark$} & 0  
            & 93.6 & 61.0 & \textbf{32.0} & 14.7 & 32.1 & 41.7 \\
          \multirow{2}{*}{} && 15 
            & 91.9 & 60.4 & 24.1 & 11.9 & 32.9 & 39.2 \\ \hline

        \end{tabular}
\end{table*}

\subsection{Datasets}
\label{sec:4.2}

To analyze how bias and unbiased streams extract features differently, we conducted experiments in videos where temporal information is crucial.
Many of the cues in videos of conventional datasets are scene information in videos, and time information is known to be not a meaningful cue.
For this reason, we used a subset of Temporal-50  \cite{Sevilla-lara_WACV2019_TemporalStaticSubset} and constructed new UCF-based subsets.

Temporal-50 is a subset with 50 classes, of which 32 are from Kinetics400 \cite{kay_arXiv2017_kinetics400} and 18 are from Something-Something \cite{Goyal_ICCV2017_SomethingSomething}.
This was created to investigate the temporal modeling of dynamic information that cannot be identified using only static cues in the scene or a single frame.
From the two datasets, the authors selected 50 classes in which temporal modeling is crucial by comparing the performances of human annotators for videos with frames in normal order versus shuffled order.
In our experiment, we use only the 32 classes of Temporal-50 which are taken from Kinetics400, and hereafter we call this ``Temporal-32''.

In addition, we constructed two datasets based on UCF101 \cite{Soomro_arXiv2012_UCF101} for this work following Temporal-50 \cite{Sevilla-lara_WACV2019_TemporalStaticSubset}. UCF-Temporal13 contains a subset of UCF101 classes that humans find difficult to recognize without viewing multiple frames. For these classes, temporal information in videos is considered important for recognition by models. In contrast, UCF-Static13 contains a class subset that humans can easily recognize even when frame order is shuffled, as the temporal information in these videos is redundant.
The class subsets are shown in Sec. \ref{sec:UCF101-subsets}.

\subsection{Evaluation metrics}
\label{sec:4.3}

We report top-1 accuracy on the validation set predicted by the unbiased stream for the model performance, and HSIC values between the two streams to show how biased and unbiased features are well separated in terms of statistical independence.

In addition, we use four metrics that can measure static bias proposed by the Human-centric Analysis Toolkit (HAT) \cite{Chung_NEURIPS2022_Action-Swap}; Background-Only Ratio (BOR), Human-Only Ratio (HOR), Swap Human Accuracy (SHAcc) and Swap Background Error (SBErr).
HAT is a tool that creates three videos from the original video;
a \emph{background-only} video where the extracted person regions are inpainted,
a \emph{human-only} video where the background (nonperson region) is masked and filled with a single color,
and a \emph{background-swapped} video where the background of the original video has been replaced with a different background (human inpainted) of videos from a different class.

BOR and HOR are the top-1 accuracy of background-only and human-only videos, normalized by the top-1 accuracy of the original video. A higher value indicates that the prediction is more dependent on background or foreground (person) information.
SHAcc is the top-1 accuracy on videos with swapped background, and a higher value indicates that the prediction is based on human information rather than the background.
SBErr represents the percentage of cases where, when the prediction is wrong for videos with swapped backgrounds, the model predicted the action class of the video from which the background is taken,
and a higher value indicates that the prediction relies more on the background information rather than foreground human actions.

\subsection{Models and settings}
\label{sec:4.1}

In the experiment, we used the TimeSformer \cite{Bertasius_ICML2021_TimeSformer} pre-trained on Kinetics400
for the unbiased stream
that encodes a video with spatio-temporal attention to extract the unbiased feature $f_u$. The same TimeSformer model is also used as the encoder of the input-based biased stream (Fig.\ref{fig:method2_DiffInput_GRL}).
For the encoder of the encoder-based biased stream for $f_b$ (Fig.\ref{fig:method2_DiffArch_GRL}), we used a late-fusion ViT \cite{Dosovitskiy_ICLR2021_ViT_Vision_transformer}, pre-trained on Kinetics400, which is applied to each frame of the input video followed by a temporal average.
However, videos were alternatively used as a frame-order shuffled version and a single-frame duplicate version, similar to the approach used for the input-based biased stream.
We compare different types of biased streams and evaluate the presence or absence of scene prediction. Note that the architecture without scene prediction is similar to CNN-based CED \cite{Bao_ICCV2021_CED} but extractors are Transformer-based, and the architecture without both biased stream and scene prediction is simply a single stream of TimeSformer.

For training with 30 epochs, we used AdamW \cite{Loshchilov_ICLR2019_AdamW} with $\beta_{1}=0.9$ and $\beta_{2}=0.98$, the
learning rate was set to 1e-5, the weight decay to 5e-5.  
For the input of the model, eight frames were sampled from a single video.
The batch size was 17 when the extractor-based biased stream was used
and 13 for the case with the input-based biased stream.
The loss hyperparameters were set to
$\alpha = 0.5$, $\beta = 0.1$, and $\lambda = 1000$.
The scheduled epoch $t_0$ was set to either 0 or 15.

\subsection{Results}
\label{sec:4.5}

Table \ref{tab:Temporal32_result} shows the results of the proposed method with different settings.
Our method with the scene prediction loss shows a smaller BOR and SBErr, in particular with
the input-based biased stream, although the best top-1 performance is with the extractor-based biased stream. This observation indicates that the proposed approach effectively reduces the static bias
related to the background of the scene.

Despite this, SHAcc is larger without scene prediction loss,
but values are small and relatively similar for all cases and need further investigation.

In contrast, HOR becomes worse when the scene prediction loss is used. This might be due to the domain gap of the human-only videos used to compute HOR.
The heads were pre-trained on normal scene images, and it does not predict soft labels reasonably for human-only videos with a masked background with a single color.
Therefore, HOR may not be a suitable measure for evaluating bias correctly,
and in the future, we will explore an appropriate measure for evaluating human-only accuracy.

Small HSIC values show how much $f_u$ and $f_u$ are independent. It is smaller for the extractor-based biased stream, and this is reasonable because encoders in biased and unbiased streams are different, while the two encoders are the same for the input-based biased stream.
The lowest HSIC was achieved with the extractor-based biased stream without scene prediction loss; however, it shows the largest BOR. Therefore,
smaller HSIC does not directly reflect smaller static bias while it is useful for disentangling biased and unbiased features.

Table \ref{tab:UCF101_result} shows the results for UCF101. Unlike Temporal32, classes of UCF101 are not considered the importance of temporal information, resulting in top-1 performance that remains almost unchanged. As expected, when the biased stream is not used, BOR is very high while HOR is low. Tables \ref{tab:UCF-temporal13_result} and \ref{tab:UCF-static13_result} show the results using UCF-Temporal13, which contains classes where temporal information appears important, and UCF-Static13, which contains classes where static information is predominant.

In both cases, BOR, HOR, and top-1 performances have improved compared to the baseline (i.e., TimeSformer which uses neither biased stream nor scene prediction loss), demonstrating the effectiveness of our proposed method for videos with extreme presence or absence of temporal information. In particular, for UCF-Temporal13, HOR values have improved significantly through the use of scene prediction loss. As the effect of scene prediction loss appears to depend heavily on the dataset and settings, we plan to conduct further verification in the future.

\section{Conclusion}

In this paper, with the aim of reducing static bias in action recognition, we proposed a method that disentangles temporal dynamic information from static scene information by leveraging the loss of statistical independence between biased and unbiased streams, along with a scene prediction loss.

Through experiments, we observed that the scene prediction loss is effective in reducing the static bias by showing a better BOR,
although it does not show consistent effectiveness in terms of the HOR metric due to the domain gap when computing HOR.

Future work includes introducing statistical independence metrics other than HSIC, and adding constraints similar to the scene prediction loss but robust to the scene appearance change.

%% file: appendix.tex
\appendix

\section{Two subsets of UCF101 classes}
\label{sec:UCF101-subsets}

\subsection{UCF-Temporal13}

``ApplyEyeMakeup'', ``ApplyLipstick'', ``Basketball'', ``BasketballDunk'', ``BodyWeightSquats'', ``BreastStroke'', ``CleanAndJerk'', ``CricketBowling'', ``CricketShot'', ``FrontCrawl'', ``HandstandPushups'', ``HandstandWalking'', ``Lunges''

\subsection{UCF-Static13}

``BaseballPitch'', ``BasketballDunk'', ``Billiards'', ``BlowingCandles'', ``FrontCrawl'', ``GolfSwing'', ``HammerThrow'', ``PoleVault'', ``Skiing'', ``SkyDiving'', ``SoccerPenalty'', ``Surfing'', ``Typing''

%% file: cameraready_iccv.bbl
\begin{thebibliography}{32}
\providecommand{\natexlab}[1]{#1}
\providecommand{\url}[1]{\texttt{#1}}
\expandafter\ifx\csname urlstyle\endcsname\relax
  \providecommand{\doi}[1]{doi: #1}\else
  \providecommand{\doi}{doi: \begingroup \urlstyle{rm}\Url}\fi

\bibitem[Bae et~al.(2025)Bae, Ahn, Kim, and Choi]{Bae_ECCV2024_Devias}
Kyungho Bae, Geo Ahn, Youngrae Kim, and Jinwoo Choi.
\newblock Devias: Learning disentangled video representations of action and scene.
\newblock In \emph{Computer Vision -- ECCV 2024}, pages 431--448, Cham, 2025. Springer Nature Switzerland.

\bibitem[Bahng et~al.(2020)Bahng, Chun, Yun, Choo, and Oh]{Bahng_ICML2020_Rebias}
Hyojin Bahng, Sanghyuk Chun, Sangdoo Yun, Jaegul Choo, and Seong~Joon Oh.
\newblock Learning de-biased representations with biased representations, 2020.

\bibitem[Bao et~al.(2021)Bao, Yu, and Kong]{Bao_ICCV2021_CED}
Wentao Bao, Qi Yu, and Yu Kong.
\newblock Evidential {Deep} {Learning} for {Open} {Set} {Action} {Recognition}.
\newblock pages 13349--13358, 2021.

\bibitem[Bertasius et~al.(2021)Bertasius, Wang, and Torresani]{Bertasius_ICML2021_TimeSformer}
Gedas Bertasius, Heng Wang, and Lorenzo Torresani.
\newblock Is space-time attention all you need for video understanding?
\newblock In \emph{Proceedings of the 38th International Conference on Machine Learning}, pages 813--824. PMLR, 2021.

\bibitem[Choi et~al.(2019)Choi, Gao, Messou, and Huang]{Choi_NEURIPS2019_Dance-in-the-mall}
Jinwoo Choi, Chen Gao, Joseph C.~E. Messou, and Jia-Bin Huang.
\newblock Why can\textquotesingle t i dance in the mall? learning to mitigate scene bias in action recognition.
\newblock In \emph{Advances in Neural Information Processing Systems}. Curran Associates, Inc., 2019.

\bibitem[Chung et~al.(2022)Chung, Wu, and Russakovsky]{Chung_NEURIPS2022_Action-Swap}
Jihoon Chung, Yu Wu, and Olga Russakovsky.
\newblock Enabling detailed action recognition evaluation through video dataset augmentation.
\newblock In \emph{Advances in Neural Information Processing Systems}, pages 39020--39033. Curran Associates, Inc., 2022.

\bibitem[Dosovitskiy et~al.(2021)Dosovitskiy, Beyer, Kolesnikov, Weissenborn, Zhai, Unterthiner, Dehghani, Minderer, Heigold, Gelly, Uszkoreit, and Houlsby]{Dosovitskiy_ICLR2021_ViT_Vision_transformer}
Alexey Dosovitskiy, Lucas Beyer, Alexander Kolesnikov, Dirk Weissenborn, Xiaohua Zhai, Thomas Unterthiner, Mostafa Dehghani, Matthias Minderer, Georg Heigold, Sylvain Gelly, Jakob Uszkoreit, and Neil Houlsby.
\newblock An image is worth 16x16 words: Transformers for image recognition at scale.
\newblock In \emph{International Conference on Learning Representations}, 2021.

\bibitem[Duan et~al.(2023)Duan, Zhao, Chen, Xiong, and Lin]{Duan_ECCV2022_Omnidebias}
Haodong Duan, Yue Zhao, Kai Chen, Yuanjun Xiong, and Dahua Lin.
\newblock Mitigating representation bias in action recognition: Algorithms and benchmarks, 2023.

\bibitem[Fioresi et~al.(2025)Fioresi, Dave, and Shah]{fioresi_ICLR2025_ALBER}
Joseph Fioresi, Ishan~Rajendrakumar Dave, and Mubarak Shah.
\newblock {ALBAR}: Adversarial learning approach to mitigate biases in action recognition.
\newblock In \emph{The Thirteenth International Conference on Learning Representations}, 2025.

\bibitem[Fukuzawa et~al.(2025)Fukuzawa, Hara, Kataoka, and Tamaki]{Fukuzawa_MMM2025_Zero_shot}
Takumi Fukuzawa, Kensho Hara, Hirokatsu Kataoka, and Toru Tamaki.
\newblock Can masking background and object reduce static bias for zero-shot action recognition?
\newblock In \emph{MMM2025}, 2025.

\bibitem[Ganin et~al.(2016)Ganin, Ustinova, Ajakan, Germain, Larochelle, Laviolette, March, and Lempitsky]{Ganin_jMLR2016_GRL}
Yaroslav Ganin, Evgeniya Ustinova, Hana Ajakan, Pascal Germain, Hugo Larochelle, Fran{\c{c}}ois Laviolette, Mario March, and Victor Lempitsky.
\newblock Domain-adversarial training of neural networks.
\newblock \emph{Journal of Machine Learning Research}, 17\penalty0 (59):\penalty0 1--35, 2016.

\bibitem[Goyal et~al.(2017)Goyal, Ebrahimi~Kahou, Michalski, Materzynska, Westphal, Kim, Haenel, Fruend, Yianilos, Mueller-Freitag, Hoppe, Thurau, Bax, and Memisevic]{Goyal_ICCV2017_SomethingSomething}
Raghav Goyal, Samira Ebrahimi~Kahou, Vincent Michalski, Joanna Materzynska, Susanne Westphal, Heuna Kim, Valentin Haenel, Ingo Fruend, Peter Yianilos, Moritz Mueller-Freitag, Florian Hoppe, Christian Thurau, Ingo Bax, and Roland Memisevic.
\newblock The "something something" video database for learning and evaluating visual common sense.
\newblock In \emph{Proceedings of the IEEE International Conference on Computer Vision (ICCV)}, 2017.

\bibitem[HARA(2020)]{Hara_IEICE2020_ActionRecognition}
Kensho HARA.
\newblock Recent advances in video action recognition with 3d convolutions.
\newblock \emph{IEICE Transactions on Fundamentals of Electronics, Communications and Computer Sciences}, advpub:\penalty0 2020IMP0012, 2020.

\bibitem[Kay et~al.(2017)Kay, Carreira, Simonyan, Zhang, Hillier, Vijayanarasimhan, Viola, Green, Back, Natsev, Suleyman, and Zisserman]{kay_arXiv2017_kinetics400}
Will Kay, Jo{\~{a}}o Carreira, Karen Simonyan, Brian Zhang, Chloe Hillier, Sudheendra Vijayanarasimhan, Fabio Viola, Tim Green, Trevor Back, Paul Natsev, Mustafa Suleyman, and Andrew Zisserman.
\newblock The kinetics human action video dataset.
\newblock \emph{CoRR}, abs/1705.06950, 2017.

\bibitem[Kim et~al.(2022)Kim, Mishra, Jin, Panda, Kuehne, Karlinsky, Saligrama, Saenko, Oliva, and Feris]{Kim_NEURIPS2022_SynAPT}
Yo-whan Kim, Samarth Mishra, SouYoung Jin, Rameswar Panda, Hilde Kuehne, Leonid Karlinsky, Venkatesh Saligrama, Kate Saenko, Aude Oliva, and Rogerio Feris.
\newblock How transferable are video representations based on synthetic data?
\newblock In \emph{Advances in Neural Information Processing Systems}, pages 35710--35723. Curran Associates, Inc., 2022.

\bibitem[Kimata et~al.(2022)Kimata, Nitta, and Tamaki]{Kimata_MMAsia2022_ObjectMix}
Jun Kimata, Tomoya Nitta, and Toru Tamaki.
\newblock Objectmix: Data augmentation by copy-pasting objects in videos for action recognition.
\newblock In \emph{Proceedings of the 4th ACM International Conference on Multimedia in Asia}, New York, NY, USA, 2022. Association for Computing Machinery.

\bibitem[Kong and Fu(2022)]{Kong_arXiv2022_ActionRecognitionPredictionSurvey}
Yu Kong and Yun Fu.
\newblock Human action recognition and prediction: A survey, 2022.

\bibitem[Lei et~al.(2023)Lei, Berg, and Bansal]{Lei_ACL2023_RevealingSingleFrameBias}
Jie Lei, Tamara Berg, and Mohit Bansal.
\newblock Revealing single frame bias for video-and-language learning.
\newblock In \emph{Proceedings of the 61st Annual Meeting of the Association for Computational Linguistics (Volume 1: Long Papers)}, pages 487--507, Toronto, Canada, 2023. Association for Computational Linguistics.

\bibitem[Li et~al.(2023)Li, Liu, Zhang, and Li]{Li_ICCV2023_StillMix}
Haoxin Li, Yuan Liu, Hanwang Zhang, and Boyang Li.
\newblock Mitigating and evaluating static bias of action representations in the background and the foreground.
\newblock In \emph{Proceedings of the IEEE/CVF International Conference on Computer Vision (ICCV)}, pages 19911--19923, 2023.

\bibitem[Li et~al.(2024)Li, Wang, He, Li, Wang, Liu, Wang, Xu, Chen, Luo, Wang, and Qiao]{Li_CVPR2024_MVBench}
Kunchang Li, Yali Wang, Yinan He, Yizhuo Li, Yi Wang, Yi Liu, Zun Wang, Jilan Xu, Guo Chen, Ping Luo, Limin Wang, and Yu Qiao.
\newblock Mvbench: A comprehensive multi-modal video understanding benchmark.
\newblock In \emph{Proceedings of the IEEE/CVF Conference on Computer Vision and Pattern Recognition (CVPR)}, pages 22195--22206, 2024.

\bibitem[Li et~al.(2018)Li, Li, and Vasconcelos]{Li_ECCV2018_Resound}
Yingwei Li, Yi Li, and Nuno Vasconcelos.
\newblock Resound: Towards action recognition without representation bias.
\newblock In \emph{Proceedings of the European Conference on Computer Vision (ECCV)}, 2018.

\bibitem[Liu et~al.(2021)Liu, Li, Niu, Xu, and Zhang]{Liu_AAAI2021_MAP-IVR}
Liu Liu, Jiangtong Li, Li Niu, Ruicong Xu, and Liqing Zhang.
\newblock Activity image-to-video retrieval by disentangling appearance and motion.
\newblock \emph{Proceedings of the AAAI Conference on Artificial Intelligence}, 35\penalty0 (3):\penalty0 2145--2153, 2021.

\bibitem[L{\'{o}}pez{-}Cifuentes et~al.(2019)L{\'{o}}pez{-}Cifuentes, Escudero{-}Vi{\~{n}}olo, Besc{\'{o}}s, and Garc{\'{\i}}a{-}Mart{\'{\i}}n]{Lopez-Cifuentes_arXiv2019_Place365}
Alejandro L{\'{o}}pez{-}Cifuentes, Marcos Escudero{-}Vi{\~{n}}olo, Jes{\'{u}}s Besc{\'{o}}s, and {\'{A}}lvaro Garc{\'{\i}}a{-}Mart{\'{\i}}n.
\newblock Semantic-aware scene recognition.
\newblock \emph{CoRR}, abs/1909.02410, 2019.

\bibitem[Loshchilov and Hutter(2019)]{Loshchilov_ICLR2019_AdamW}
Ilya Loshchilov and Frank Hutter.
\newblock Decoupled weight decay regularization.
\newblock In \emph{International Conference on Learning Representations}, 2019.

\bibitem[Sevilla-Lara et~al.(2019)Sevilla-Lara, Zha, Yan, Goswami, Feiszli, and Torresani]{Sevilla-lara_WACV2019_TemporalStaticSubset}
Laura Sevilla-Lara, Shengxin Zha, Zhicheng Yan, Vedanuj Goswami, Matt Feiszli, and Lorenzo Torresani.
\newblock Only {Time} {Can} {Tell}: {Discovering} {Temporal} {Data} for {Temporal} {Modeling}, 2019.
\newblock arXiv:1907.08340 [cs].

\bibitem[Song et~al.(2012)Song, Smola, Gretton, Bedo, and Borgwardt]{Song_JMLR2012_DependenceMaximization}
Le Song, Alex Smola, Arthur Gretton, Justin Bedo, and Karsten Borgwardt.
\newblock Feature {Selection} via {Dependence} {Maximization}.
\newblock \emph{Journal of Machine Learning Research}, 13\penalty0 (47):\penalty0 1393--1434, 2012.

\bibitem[Soomro et~al.(2012)Soomro, Zamir, and Shah]{Soomro_arXiv2012_UCF101}
Khurram Soomro, Amir~Roshan Zamir, and Mubarak Shah.
\newblock {UCF101:} {A} dataset of 101 human actions classes from videos in the wild.
\newblock \emph{CoRR}, abs/1212.0402, 2012.

\bibitem[Sugiura and Tamaki(2024)]{Sugiura_VISAPP2024_S3Aug}
Taiki Sugiura and Toru Tamaki.
\newblock S3aug: Segmentation, sampling, and shift for action recognition.
\newblock In \emph{Proceedings of the 19th International Joint Conference on Computer Vision, Imaging and Computer Graphics Theory and Applications - Volume 2: VISAPP}, pages 71--79. INSTICC, SciTePress, 2024.

\bibitem[Ulhaq et~al.(2022)Ulhaq, Akhtar, Pogrebna, and Mian]{Ulhaq_arXiv2022_VisionTransformersActionRecognitionSurvey}
Anwaar Ulhaq, Naveed Akhtar, Ganna Pogrebna, and Ajmal Mian.
\newblock Vision transformers for action recognition: A survey, 2022.

\bibitem[Wang et~al.(2021)Wang, Gao, Li, Lin, Ma, Cheng, Peng, Huang, Ji, and Sun]{Wang_CVPR2021_RemovingBackground}
Jinpeng Wang, Yuting Gao, Ke Li, Yiqi Lin, Andy~J. Ma, Hao Cheng, Pai Peng, Feiyue Huang, Rongrong Ji, and Xing Sun.
\newblock Removing the background by adding the background: Towards background robust self-supervised video representation learning.
\newblock In \emph{Proceedings of the IEEE/CVF Conference on Computer Vision and Pattern Recognition (CVPR)}, pages 11804--11813, 2021.

\bibitem[Yun et~al.(2022)Yun, Kim, Han, Song, Ha, and Shin]{Yun_PMLR2022_TIME}
Sukmin Yun, Jaehyung Kim, Dongyoon Han, Hwanjun Song, Jung-Woo Ha, and Jinwoo Shin.
\newblock Time {Is} {MattEr}: {Temporal} {Self}-supervision for {Video} {Transformers}.
\newblock In \emph{Proceedings of the 39th {International} {Conference} on {Machine} {Learning}}, pages 25804--25816. PMLR, 2022.
\newblock ISSN: 2640-3498.

\bibitem[Zhong et~al.(2023)Zhong, Mishra, Kim, Jin, Panda, Kuehne, Karlinsky, Saligrama, Oliva, and Feris]{Zhong_NEURIPS2023_PPMA}
Howard Zhong, Samarth Mishra, Donghyun Kim, SouYoung Jin, Rameswar Panda, Hilde Kuehne, Leonid Karlinsky, Venkatesh Saligrama, Aude Oliva, and Rogerio Feris.
\newblock Learning human action recognition representations without real humans.
\newblock In \emph{Advances in Neural Information Processing Systems}, pages 65069--65087. Curran Associates, Inc., 2023.

\end{thebibliography}
